\documentclass{article}
\usepackage{marvosym}

\PassOptionsToPackage{numbers, compress}{natbib}
\usepackage[preprint]{neurips_2025}

\usepackage[utf8]{inputenc} % allow utf-8 input
\usepackage[T1]{fontenc}    % use 8-bit T1 fonts
\usepackage{hyperref}       % hyperlinks
\usepackage{url}
\usepackage{makecell}
\usepackage{graphicx}
\usepackage{amsmath}
\usepackage{amssymb}
\usepackage{booktabs}
\usepackage{comment}
\usepackage{threeparttable}
\usepackage{xcolor}
\usepackage{multirow}
\usepackage{verbatimbox}
\usepackage{subcaption}
\usepackage{tcolorbox}
\newcounter{subtable@save}
\usepackage[export]{adjustbox}
\definecolor{darkgray}{gray}{0.7}

\definecolor{clinicalgreen}{HTML}{BDD0BA}
\definecolor{answerbox}{RGB}{220,240,230}
\definecolor{lightbluebox}{RGB}{230,245,255}

\title{Beyond Distillation: Pushing the Limits of Medical LLM Reasoning with Minimalist Rule-Based RL}

\author{Che Liu$^{1}$\thanks{Equal Contribution},
 Haozhe Wang$^{2*}$,
    Jiazhen Pan$^{3*}$,\\  %
 \textbf{Zhongwei Wan}$^4$\,,  %
 \textbf{Yong Dai}$^5$\,,
  \textbf{Fangzhen Lin}$^2$\,,\\
  \textbf{Wenjia Bai}$^1$\,,
  \textbf{Daniel Rueckert}$^{1,3}$\,, 
  \textbf{Rossella Arcucci}$^{1}$\\
	$^1$Imperial College London, ~$^2$HKUST, 
        $^3$Technical University of Munich , \\
        $^4$Ohio State University, ~$^5$Fudan University
 \\
 \Letter~\textit{che.liu21@imperial.ac.uk}\\
 \\
Project page: \href{https://cheliu-computation.github.io/AlphaM/}{\texttt{https://cheliu-computation.github.io/AlphaMed/}}
 }

\begin{document}

\maketitle
\begin{abstract} 
Improving performance on complex tasks and enabling interpretable decision making in large language models (LLMs), especially for clinical applications, requires effective reasoning. Yet this remains challenging without supervised fine-tuning (SFT) on costly chain-of-thought (CoT) data distilled from closed-source models (e.g., GPT-4o). In this work, we present \textbf{AlphaMed}, the first medical LLM to show that reasoning capability can emerge purely through reinforcement learning (RL), using minimalist rule-based rewards on public multiple-choice QA datasets, without relying on SFT or distilled CoT data. AlphaMed achieves state-of-the-art results on six medical QA benchmarks, outperforming models trained with conventional SFT+RL pipelines. On challenging benchmarks (e.g., MedXpert), AlphaMed even surpasses larger or closed-source models such as DeepSeek-V3-671B and Claude-3.5-Sonnet. 
To understand the factors behind this success, we conduct a comprehensive data-centric analysis guided by three questions: 
\textbf{(i)} Can minimalist rule-based RL incentivize reasoning without distilled CoT supervision? 
\textbf{(ii)} How do dataset quantity and diversity impact reasoning? 
\textbf{(iii)} How does question difficulty shape the emergence and generalization of reasoning? 
Our findings show that dataset informativeness is a key driver of reasoning performance, and that minimalist RL on informative, multiple-choice QA data is effective at inducing reasoning without CoT supervision. 
We also observe divergent trends across benchmarks, underscoring limitations in current evaluation and the need for more challenging, reasoning-oriented medical QA benchmarks. The code and pretrained model weights will be publicly released upon acceptance.
\end{abstract}

\section{Introduction}
Recently, the reasoning capabilities of large language models (LLMs) have advanced significantly, achieving impressive results in tasks requiring complex reasoning, such as mathematical problem solving, code generation, and general-purpose benchmarks~\cite{o1journey,o1p1,o1p9,guan2024deliberative}. These developments highlight the potential of LLMs to generalize and perform multi-step reasoning across domains.
In the medical domain, reasoning is particularly crucial. Clinical natural language processing (NLP) tasks often require interpreting nuanced patient information, integrating knowledge from diverse sources, and making informed decisions~\cite{saab2024capabilities,chen2024cod,patel2005thinking}. More importantly, reasoning provides a valuable lens into the model’s decision-making process, allowing researchers and clinicians to examine how conclusions are derived. This improves the interpretability and transparency of AI outputs, which are essential for clinical trust~\cite{o12,o1p10}.

Currently, most medical LLMs acquire reasoning capabilities through supervised fine-tuning (SFT) on chain-of-thought (CoT) datasets, often followed by reinforcement learning (RL) for further refinement. However, this pipeline heavily relies on an initial SFT stage using costly CoT data, which are either manually crafted or distilled from closed-source commercial models such as GPT-4o~\cite{o1medicalpreliminary,chen2023huatuogpt}. 
This dependence not only incurs substantial annotation and distillation costs but also introduces scalability and accessibility challenges, as it ties model development to expensive and external resources. These limitations motivate a critical question: 
\begin{tcolorbox}[
    colback=lightbluebox,    % light blue background
    colframe=blue!40!black,  % frame color (soft blue-gray)
    arc=8pt,                % rounded corners (all four)
    boxrule=0.8pt,           % border thickness
    left=2mm, right=2mm,     % horizontal padding
    top=1mm, bottom=1mm,     % vertical padding
]
\centering
\textbf{Can we achieve medical reasoning through minimalist rule-based RL\vspace{1mm}\\
\textbf{\large without relying on distilled CoT data?}}
\end{tcolorbox}

To address this question, we propose \textbf{AlphaMed}, the first work designed to incentivize reasoning capability solely through minimalist rule-based RL, going beyond conventional approaches that rely on SFT with CoT data. Instead of depending on distilled CoT data supervision, AlphaMed is trained directly via simple rule-based rewards derived from multiple-choice QA datasets. Our key contributions are as follows:
\begin{itemize}
    \item We show that minimalist rule-based RL can incentivize reasoning ability in medical LLMs without relying on distilled CoT data, achieving superior performance. We further analyze how dataset quantity, diversity, and especially informativeness impact reasoning performance. We empirically find that higher informativeness enhances reasoning performance, while less-informative data limits gains.
    
    \item We show that reasoning can be incentivized even with lower-difficulty data and further enhanced by harder examples. While high-difficulty samples benefit challenging benchmarks like MedXpert, a mix of difficulty levels is essential for robust generalization. Non-monotonic trends across benchmarks suggest that current evaluations may be insufficient to assess medical LLM reasoning.

    \item Building on these insights, we introduce AlphaMed, a medical LLM trained solely via minimalist rule-based RL without any SFT on distilled CoT data, and demonstrate that it achieves state-of-the-art performance across six mainstream medical QA benchmarks, outperforming models that use complex training strategies with CoT data and even surpassing larger or closed-source models such as DeepSeek-V3-671B and GPT-4o.
    
\end{itemize}

\section{Related Work}

\paragraph{Supervised Fine-Tuning for Reasoning in LLMs.}
Large language models can acquire complex reasoning skills through SFT on CoT data. For example, \cite{wei2022chainofthought} showed that training models to generate step-by-step reasoning paths significantly improves performance on math and logic problems. \cite{chung2022scaling} scaled this approach by incorporating a broad range of CoT examples into instruction tuning across diverse tasks. \cite{zelikman2022star} proposed STaR, where a model bootstraps its own reasoning traces to reduce reliance on human-annotated CoT. However, recent work \cite{chu2025sft} suggests that SFT often encourages memorization of training rationales rather than true reasoning generalization, limiting robustness in out-of-distribution or unfamiliar tasks. Moreover, obtaining high-quality CoT data is costly, requiring either expert annotations or distillation from proprietary models, posing significant challenges to scalability and adaptability~\cite{vl-rethinker}.

\paragraph{Reinforcement Learning with Preference Data after SFT.}
InstructGPT \cite{ouyang2022training} introduced reinforcement learning with human preferences (RLHF) to align model behavior with user intent. Subsequent research has shown that RL can enhance generalization~\cite{vl-rethinker, wang2025learning} and better capture nuanced human preferences beyond rote memorization \cite{chu2025sft}. Among RL algorithms, Proximal Policy Optimization (PPO) is widely used, but it is highly resource-intensive—requiring learned reward models that are often sensitive to noise, difficult to interpret, and occasionally misaligned with intended objectives \cite{rafailov2023direct}. To address these limitations, Direct Preference Optimization (DPO) \cite{rafailov2023direct} eliminates the need for an explicit reward model by directly optimizing over preference pairs. However, DPO still relies on high-quality preference annotations, which are particularly challenging to construct in the medical domain due to clinical ambiguity and a lack of universal agreement on what constitutes a ``better'' response \cite{ura2024openbiollm}. Recently, DeepSeek-R1-Zero~\cite{guo2025deepseek} demonstrated that reasoning behavior can be effectively elicited without CoT supervision or preference annotations, instead by leveraging final answers (e.g., multiple-choice accuracy) as rule-based supervision signals~\cite{vl-rethinker, wang2025learning, zeng2025acecoder, pan2025medvlm}.

\paragraph{Open-Source Medical LLMs.}
Open-source medical LLMs have emerged as promising tools for domain-specific clinical reasoning, yet most remain heavily dependent on supervised data or handcrafted feedback. HuatuoGPT \cite{zhang2023huatuogpt} was instruction-tuned on ChatGPT-distilled medical dialogues. BioMistral \cite{labrak2024biomistral} adapted the Mistral architecture to biomedical question answering through continued pretraining~\cite{ke2023continual} and domain-specific instruction tuning. OpenBioLLM \cite{ura2024openbiollm} and UltraMedical \cite{zhang2024ultramedical} utilized DPO-based preference optimization, but their preference pairs were directly distilled from closed-source models, making supervision ambiguous and potentially inconsistent with expert clinical reasoning. Since human verification of each distilled example is prohibitively costly and impractical, there is no guarantee that the reasoning process reflected in the supervision is valid.
HuatuoGPT-o1 \cite{chen2024huatuogpt-o1} further incorporated PPO using a self-trained 3B reward model and relied on CoT data distilled from OpenAI o1. However, this approach is resource-intensive and tightly coupled to the quality and coverage of proprietary data, limiting its scalability and generalizability. m1~\cite{huang2025m1} also adopts SFT on distilled chain-of-thought data, where step-by-step reasoning traces are generated by external large reasoning model, thus still relying on distilled CoT data.

\section{Preliminaries}

\paragraph{Group Relative Policy Optimization (GRPO)}
Given a question-answer pair \((q, a)\), the behaviour policy \(\pi_{\text{old}}\) generates a set of \(G\) candidate completions \(\{o_i\}_{i=1}^G\) for each question \(q\). Each response receives a scalar reward \(r_i\), which may be derived from human preference comparisons or automated scoring heuristics; in this work, we use a rule-based reward. The relative quality of each response is assessed within the group through normalization. The training objective is:
\vspace{-5pt}
\begin{equation}
\begin{aligned}
 \mathcal{J}_{\text{GRPO}}(\theta) =\; & \mathbb{E}_{(q,a) \sim \mathcal{D}, \{o_i\}_{i=1}^G \sim \pi_{\text{old}}(\cdot|q)} \Bigg[ \frac{1}{G} \sum_{i=1}^{G} \frac{1}{|o_i|} \sum_{t=1}^{|o_i|} \Big( \\
& \min\left( r_{i,t}(\theta) \hat{A}_{i,t}, \text{clip}(r_{i,t}(\theta), 1-\epsilon, 1+\epsilon)\hat{A}_{i,t} \right) \Big) \Bigg]
\end{aligned}
\label{eq:grpo}
\end{equation}
where the group-normalized advantage $\hat{A}_{i,t}$ and the token-level importance weight $r_{i,t}(\theta)$ are defined as:
\[
\hat{A}_{i,t} = \frac{r_i - \text{mean}(\{r_j\}_{j=1}^G)}{\text{std}(\{r_j\}_{j=1}^G)}, \quad
r_{i,t}(\theta) = \frac{\pi_\theta(o_{i,t} \mid q, o_i, <t)}{\pi_{\text{old}}(o_{i,t} \mid q, o_i, <t)}.
\]
Here, \(\epsilon\) is a hyperparameter controlling the tolerance for policy deviation. The \texttt{clip} function prevents large updates by ensuring that the ratio between the current and reference policy stays within a predefined range. Specifically, it clips the importance weight \(r_{i,t}(\theta)\) to the interval \([1-\epsilon, 1+\epsilon]\), thereby stabilizing training and mitigating the risk of policy collapse.
This objective encourages the model to improve token probabilities for completions with above-average rewards, while stabilizing updates via a clipped importance weight similar to PPO \cite{ppo2017}.

\paragraph{Rule-based Reward Modelling}  
To enable minimalist RL without relying on external verifiers or human-provided rewards, we adopt a simple rule-based approach consistent with~\cite{guo2025deepseek}. This method directly evaluates the correctness of the model’s output using binary feedback, eliminating the need for a separate reward model:
\begin{equation}
r_i = 
\begin{cases}
1, & \text{if } \texttt{is\_answer\_correct}(\hat{y}_i, y) \\
0, & \text{otherwise}
\end{cases}
\label{eq:reward}
\end{equation}
Here, \(y\) is the ground-truth answer and \(\hat{y}_i\) denotes the model-generated prediction from the \(i\)-th output \(o_i\).  
This straightforward reward mechanism provides a clear supervision signal grounded in task accuracy. By leveraging structured outputs (e.g., multiple-choice answers), we enable effective RL without manually written rationales or preference annotations.

\section{AlphaMed}

\subsection{Training Configuration}
We aim to elicit medical reasoning behavior purely through rule-based RL, without relying on SFT with CoT data or RL with rewards from external verifiers. To ensure a fair comparison with \texttt{HuatuoGPT-o1}~\cite{chen2024huatuogpt-o1}, we adopt \texttt{Llama3.1-8B-Instruct} and \texttt{Llama3.1-70B-Instruct} as backbone models. All experiments are conducted under full parameter tuning with a batch size of 512, meaning each batch contains 64 QA pairs and each question generates 8 candidate answers, trained for 300 steps. We use \texttt{verl}\footnote{https://github.com/volcengine/verl}~\cite{sheng2024hybridflow}, a framework designed for rule-based RL. 
A simple binary reward function, defined in Eq.~\ref{eq:reward}, assigns 1 if the model's response ends with a correctly formatted boxed answer matching the ground truth (e.g., \texttt{\textbackslash boxed\{C\}}), and 0 otherwise.
The model is optimized using the GRPO objective described in Eq.~\ref{eq:grpo}. We train the 8B model on 8 Nvidia A800-80G GPUs and the 70B model on 64 A800-80G GPUs.

\subsection{Evaluation Configuration}
\textbf{Datasets.} We evaluate our models on six medical QA benchmarks, using accuracy as the evaluation metric across all datasets. These include MedQA-USMLE~\cite{jin2021disease} (MedQA), MedMCQA~\cite{pal2022medmcqa} (MedMCQA), PubMedQA~\cite{jin2019pubmedqa} (PubMedQA), MMLU-Pro medical subsets~\cite{wang2024mmlu} (MMLU-ProM), GPQA medical subsets~\cite{rein2024gpqa} (GPQA-M), and the most recent and challenging large-scale dataset, MedXpertQA~\cite{zuo2025medxpertqa} (MedXpert). Details are provided in Sec. \ref{sec:eval data}.

Based on their levels of challenge \cite{tang2025medagentsbench}, we categorize MedQA, MedMCQA, and PubMedQA~\cite{jin2021disease,pal2022medmcqa,jin2019pubmedqa} as \textit{\textbf{normal}}, while MMLU-ProM and GPQA-M~\cite{wang2024mmlupro,rein2024gpqa} are classified as \textit{\textbf{hard}}, as they primarily target advanced expert-level knowledge. Finally, MedXpert~\cite{zuo2025medxpertqa} is designated as \textit{\textbf{hard+}}, as the original work explicitly highlights its focus on complex clinical reasoning and expert-level decision making, positioning it as one of the most challenging benchmarks to date.

\textbf{Baseline Methods.} We compare against a broad range of general and medical-specific LLM baselines. General-purpose base instruct models include \texttt{Qwen2.5-7B/32B/72B} and \texttt{Llama3.1-8B/70B}. Medical-specific models cover \texttt{MedLlama3}, \texttt{OpenBioLLM}~\cite{pal2024openbiollms}, \texttt{MMed} and \texttt{MMed-S}~\cite{qiu2024towards}, \texttt{Med42}~\cite{christophe2024med42}, and \texttt{UltraMedical}~\cite{zhang2024ultramedical}, which leverage distilled preference data and RL following SFT. \texttt{HuatuoGPT-o1}~\cite{chen2024huatuogpt-o1} is trained on CoT data distilled from GPT-4o using model-based RL with a large (3B) reward model. \texttt{m1}~\cite{huang2025m1} is similarly trained with extensive CoT distilled from DeepSeekR1~\cite{guo2025deepseek} via SFT.

\section{Experiments}
\subsection{Data Curation}
\label{sec:data curation}
\noindent\textbf{Initial Data Collection.}
% \paragraph{Initial Data Collection.}
Following~\cite{huang2025m1}, we collect the training splits of three large-scale public multiple-choice medical QA datasets: MedQA~\cite{medqa}, MedMCQA~\cite{medmcqa}, and PubMedQA~\cite{jin2019pubmedqa}\footnote{We use the official training splits of all three datasets.}\footnote{For PubMedQA~\cite{jin2019pubmedqa}, only questions with definitive answer labels (i.e., A/B/C) are retained.}. MedQA~\cite{medqa} contains expert-level clinical questions from the USMLE. MedMCQA~\cite{medmcqa} includes factoid and reasoning questions from Indian medical entrance exams (AIIMS, NEET).  PubMedQA~\cite{jin2019pubmedqa} focuses on biomedical research question answering. Notably, its training split is automatically generated by a machine learning model that heuristically converts biomedical research article abstract into yes/no questions and assigns answers based on negation cues. The dataset statistics are summarized in Sec.~\ref{sec:diff dist}.
\\
\noindent\textbf{Quantifying Data Difficulty.} 
% \paragraph{Quantifying Data Difficulty.} 
To quantify question difficulty, we perform inference using \texttt{Llama3.1-8B-Instruct}~\cite{llama3}. For each question, we generate five reasoning completions with the following prompt:  
\texttt{``Please reason step by step, and put the final answer in \textbackslash boxed\{\}"}.  
We then calculate the proportion of correct predictions among the five outputs, which serves as a proxy for the question's difficulty.  Based on this proportion, we categorize questions into six difficulty levels (L1--L6). Specifically, L1 includes questions where all five completions are correct, L2 where four are correct, and so on, with L6 representing questions where all five completions are incorrect. The difficulty level distribution of each train set as shown in Tab. \ref{tab:diff dist}

\begin{figure}[t!]
    \centering
    \includegraphics[width=0.99\linewidth]{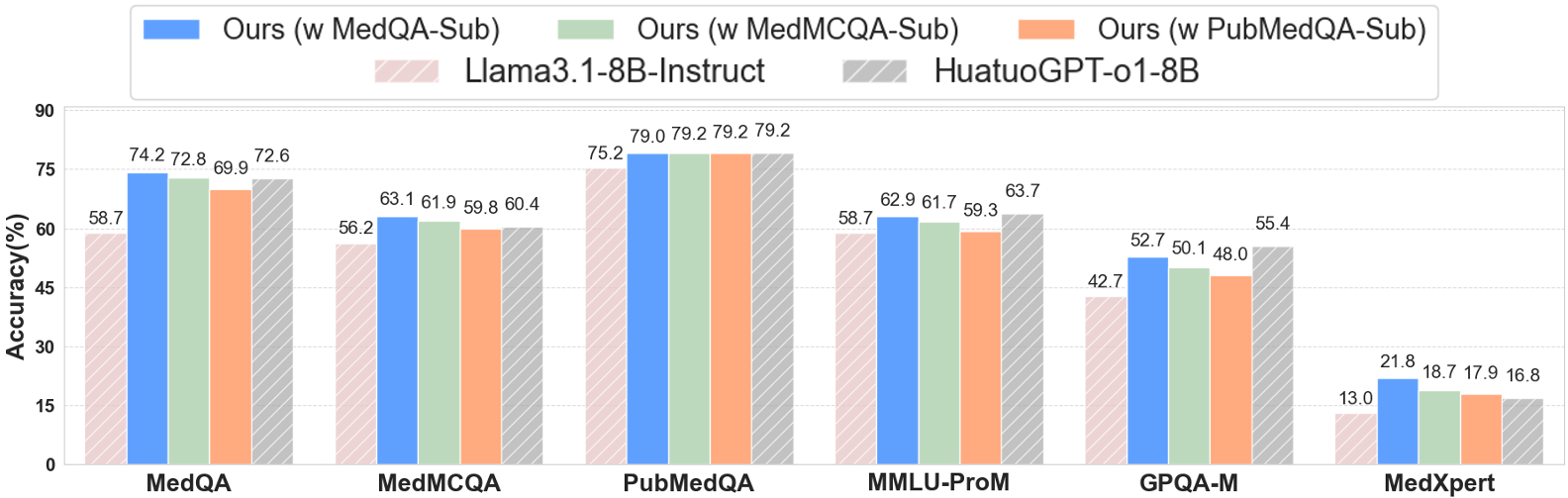}
    \vspace{-3pt}
    \caption{\textbf{Performance comparison on six medical QA benchmarks.} 
    Our models are initialized with \texttt{Llama3.1-8B-Instruct}~\cite{llama3} and trained using minimalist rule-based RL on one of three balanced subsets: \textit{MedQA-Sub}, \textit{MedMCQA-Sub}, or \textit{PubMedQA-Sub} (shown as \textcolor{blue}{blue}, \textcolor{green!50!black}{green}, and \textcolor{orange}{orange} bars, respectively). Despite using only 1,200 examples per subset, all variants of our model achieve substantial improvements over the base \texttt{Llama3.1-8B-Instruct} and match or surpass the strong baseline \texttt{HuatuoGPT-o1-8B} across all benchmarks.}
    \label{fig:single-sub-dataset}
    \vspace{-10pt}
\end{figure}

\begin{figure}[ht!]
    \centering
    \includegraphics[width=0.99\linewidth]{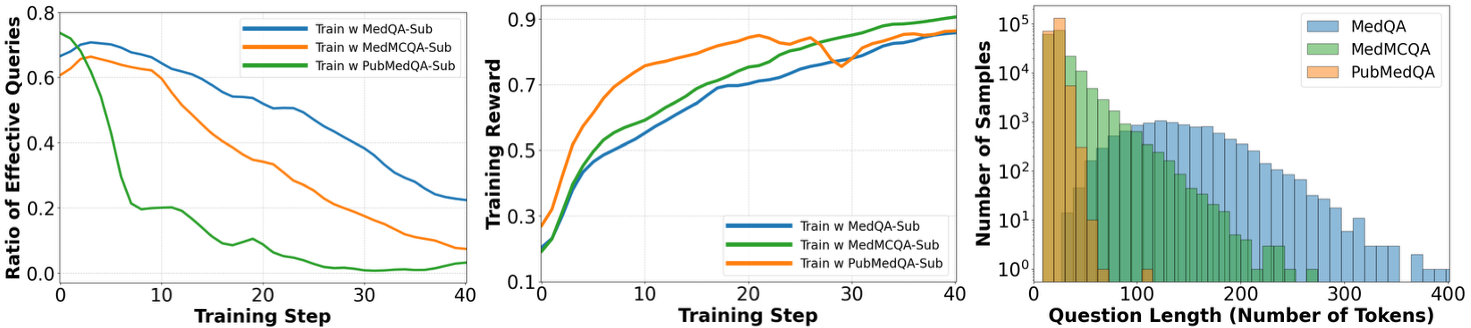}
    \caption{
    \textbf{Dataset analysis and training dynamics.}
    \textbf{Left:} Ratio of effective queries over training steps; each curve corresponds to models trained on a specific subset.
    \textbf{Middle:} Training reward per step for models trained on each subset.
    \textbf{Right:} Distribution of question lengths (number of tokens) in MedQA, MedMCQA, and PubMedQA~\cite{medqa,medmcqa,jin2019pubmedqa}.
    }
\vspace{-20pt}
    \label{fig:informative}
\end{figure}

\subsection{RQ1: Can Minimalist RL Incentivize Medical Reasoning Without Distilled-CoT SFT?}
To investigate whether minimalist rule-based RL can incentivize medical reasoning in LLMs without relying on SFT with distilled CoT data, we conduct a pilot study by sampling 200 examples from each difficulty level to construct three balanced subsets (1,200 samples each) from three public medical QA datasets: \textit{MedQA-Sub}, \textit{MedMCQA-Sub}, and \textit{PubMedQA-Sub}.  
We use \texttt{Llama3.1-8B-Instruct} as the backbone model and train it separately on each subset using minimalist RL.  
As shown in Fig.~\ref{fig:single-sub-dataset}, all models trained on these subsets achieve substantial gains over the original backbone across all six benchmarks (e.g., +15.5\% on MedQA, +8.8\% on MedXpert).  
Remarkably, all variants trained on different subsets perform comparably to or even surpass \texttt{HuatuoGPT-o1-8B}~\cite{chen2024huatuogpt}, a strong baseline trained via SFT on CoT data distilled from GPT-4o~\cite{gpt4o} and further fine-tuned with RL using a 3B reward model.  
Notably, on MedXpert~\cite{zuo2025medxpertqa}, the most challenging benchmark, all three variants outperform \texttt{HuatuoGPT-o1-8B}~\cite{chen2024huatuogpt}.  
These results demonstrate that reasoning capability can be effectively incentivized through minimalist RL on small-scale, low-cost multiple-choice QA data, without relying on SFT with distilled CoT data, and can even outperform models trained with more complex strategies.

Surprisingly, \textbf{multistep reasoning} (e.g., \texttt{Step 1..., Step 2...}; see Fig.~\ref{fig:case1}, \ref{fig:case2}, \ref{fig:case3}) spontaneously emerges in the model’s output, which derives the final answer through sequential analysis, despite being supervised only on the final choice, without intermediate reasoning traces like distilled CoT data~\cite{huang2025m1,chen2024huatuogpt}.
This emergent behavior shows that minimalist rule-based RL not only boosts performance but also encourages structured reasoning, offering valuable interpretability into the model’s decision-making.

\paragraph{Performance Variation and Training Dynamics Across Subsets.}
We observe clear performance differences among training subsets, consistently ranking as \textit{MedQA-Sub} $>$ \textit{MedMCQA-Sub} $>$ \textit{PubMedQA-Sub}. To understand this variation, we explore the training dynamics of models trained on each subset. As depicted in Fig.~\ref{fig:informative} (left), following~\cite{vl-rethinker}, the ratio of effective queries is computed as $1 - \frac{\# \text{solved all} + \# \text{solved none}}{\# \text{unique queries}}$, where ``solved all'' and ``solved none'' denote batches in which all responses are either correct or incorrect. Models trained on \textit{PubMedQA-Sub} exhibit a rapid decline in the effective query ratio, indicating premature saturation and a reduction in effective samples from the batch. The training reward in Fig.~\ref{fig:informative} (middle) further supports this: the \textit{PubMedQA-Sub} variant starts with a higher initial reward and increases rapidly, suggesting that the data is easy to learn at the start, but quickly saturates after about 20 steps. In contrast, the \textit{MedQA-Sub} and \textit{MedMCQA-Sub} models improve steadily throughout training.

\paragraph{Dataset Informativeness as a Key Driver.}
To further investigate these dynamics, we analyze the question length distributions in the source datasets of each subset, as shown in Fig.~\ref{fig:informative} (right). Notably, MedQA~\cite{medqa} exhibits a significantly longer question length distribution compared to MedMCQA~\cite{medmcqa} and PubMedQA \cite{jin2019pubmedqa}, this ordering closely matches the observed performance of model variants trained on the respective subsets. These differences are linked to dataset construction mechanisms: PubMedQA  \cite{jin2019pubmedqa} is automatically curated from biomedical literature, often resulting in noisier and less informative questions; MedMCQA \cite{medmcqa} is based on human-authored medical school entrance exams, providing more reliable and informative samples; MedQA \cite{medqa} is sourced from the USMLE, a challenging licensing exam, and thus contains the most informative and well-structured questions. Altogether, our findings suggest that question length serves as a practical proxy for dataset informativeness in medical QA. High-informativeness, exam-certified data provide more stable and effective learning signals for minimalist RL, whereas noisy, automatically curated data may offer lower informativeness and thus hinder the acquisition of reasoning ability.

\vspace{-5pt}
\begin{tcolorbox}[
    colback=answerbox,
    colframe=teal!60!black,
    arc=6pt,
    boxrule=0.8pt,
    left=2mm, right=2mm,
    top=1mm, bottom=1mm,
]
\textbf{Finding 1.1:}
Minimalist rule-based RL enables medical reasoning in LLM beyond reliance on SFT with distilled CoT data. \vspace{1mm}\\
\noindent\textbf{Finding 1.2:}
Dataset informativeness is critical for training success. LLM trained on low informative or noisy data exhibit degraded performance. Question length serves as a practical proxy for informativeness in medical QA.
\end{tcolorbox}
\vspace{-10pt}
\subsection{RQ2: Impact of Dataset Quantity and Diversity}
\begin{figure}[t!]
\vspace{0pt}
    \centering
    % First row: quantity
    \begin{minipage}[c]{0.49\linewidth}
    \vspace{0pt}
        \includegraphics[width=\linewidth]{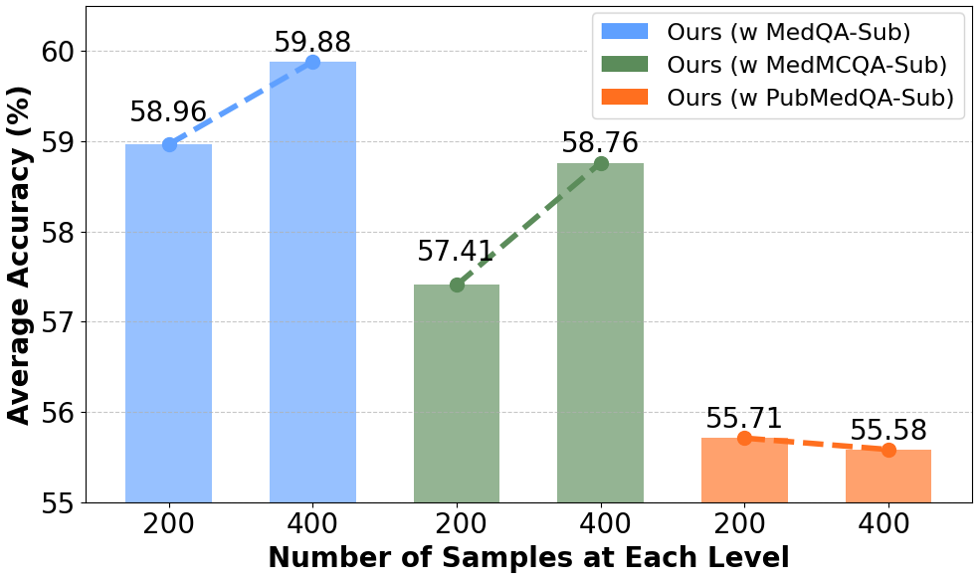}
    \end{minipage}
    \hfill
    \begin{minipage}[c]{0.49\linewidth}
    \caption{
    \textbf{Effect of data quantity.}
    Average accuracy across six medical QA benchmarks as the number of samples per level increases from 200 to 400, resulting in the total subset size growing from 1,200 to 2,400 examples. Scaling \textit{MedQA-Sub} and \textit{MedMCQA-Sub} leads to consistent performance gains, highlighting the value of informative data. In contrast, \textit{PubMedQA-Sub} shows no improvement, reflecting the limitations of low-informative data sources.
    }
    \label{fig:quantity}
    \end{minipage}
    \\
    % Second row: diversity
    \begin{minipage}[c]{0.53\linewidth}
    \vspace{0pt}
        \includegraphics[width=\linewidth]{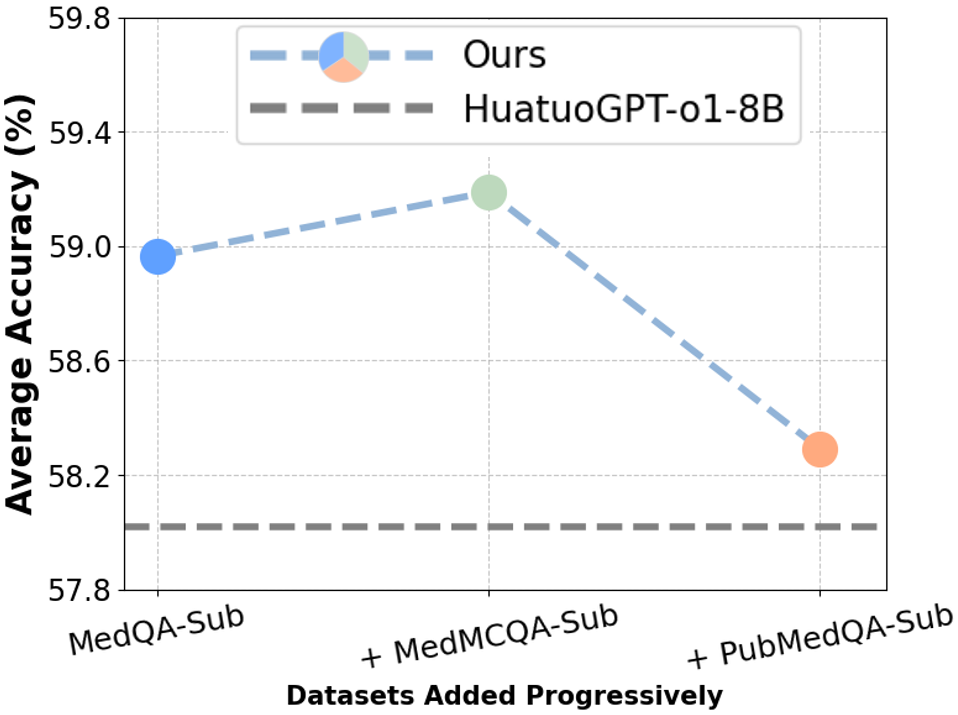}
    \end{minipage}
    \hfill
    \begin{minipage}[c]{0.46\linewidth}
    \caption{
    \textbf{Effect of data diversity.}
    Average accuracy across six medical QA benchmarks when models are trained individually on single or combined subsets. Adding \textit{MedMCQA-Sub} to \textit{MedQA-Sub} boosts performance, while further adding \textit{PubMedQA-Sub} reduces it, suggesting that less informative data can negate the benefits of increased diversity.
    }
        \label{fig:diversity}
    \end{minipage}
    \vspace{-20pt}
\end{figure}
\paragraph{Effect of Dataset Quantity.}
To investigate the effect of training data size, we increase the number of samples per difficulty level from 200 to 400 for each of the three subsets, resulting in the total number of samples in each subset increasing from 1,200 to 2,400. As shown in Fig.~\ref{fig:quantity}, we report the average accuracy across six benchmarks. Scaling \textit{MedQA-Sub} improves accuracy from 58.96\% to 59.88\%, and \textit{MedMCQA-Sub} improves from 57.41\% to 58.76\%, demonstrating that increasing high-informative data benefits model performance.  
In contrast, scaling \textit{PubMedQA-Sub} yields no improvement (55.71\% → 55.58\%), suggesting that adding more low-informative or noisy samples may degrade performance rather than enhance it.

\paragraph{Effect of Dataset Diversity.}
We further examine the effect of dataset diversity by progressively combining subsets. As shown in Fig.~\ref{fig:diversity}, adding \textit{MedMCQA-Sub} to \textit{MedQA-Sub} further improves performance, highlighting the benefit of combining diverse and informative datasets. However, incorporating \textit{PubMedQA-Sub} reverses the upward trend and leads to a decline in performance, indicating that noisy and less informative data not only fail to contribute but may also harm reasoning ability.

\vspace{-5pt}
\begin{tcolorbox}[
    colback=answerbox,      % background color
    colframe=teal!60!black, % border color
    arc=6pt,    % rounded corners, like your example
    boxrule=0.8pt,          % border thickness
    left=2mm, right=2mm,    % padding
    top=1mm, bottom=1mm,
]
\textbf{Finding 2:}
Performance improves with increased data quantity and diversity only when the additional samples are informative; low-quality data harms the learning of reasoning ability.
\end{tcolorbox}
\vspace{-10pt}
\begin{figure}[t!]
    \centering
    \includegraphics[width=0.99\linewidth]{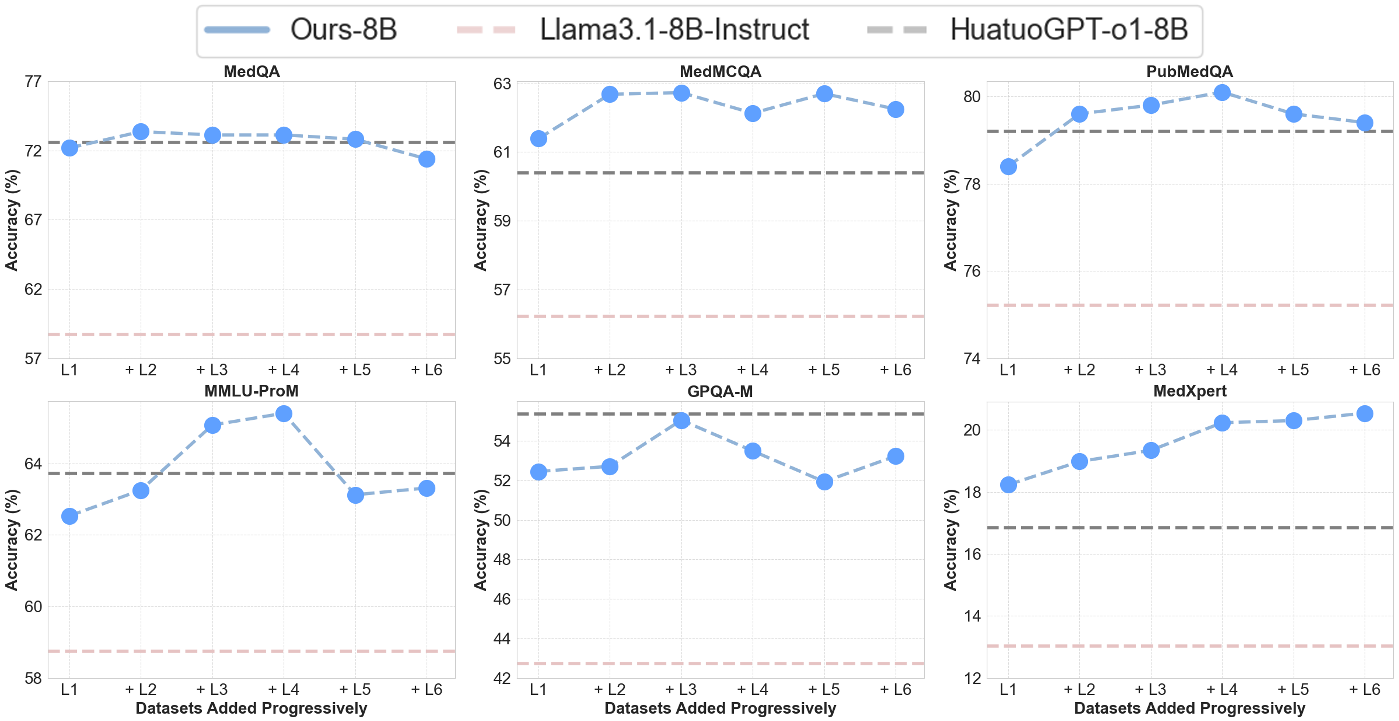}
    \vspace{-5pt}
    \caption{Performance on six benchmarks when training on subsets with increasing difficulty levels (L1 to L6). Each blue dot represents a separately trained model on a subset that includes all data up to the indicated difficulty level; new data are incorporated only through separate training runs, not incrementally during training. While performance on MedXpert~\cite{zuo2025medxpertqa} increases consistently, trends on other benchmarks vary. Final models trained on the full set (L1--L6) generally achieve comparable or superior performance to HuatuoGPT-o1-8B~\cite{chen2024huatuogpt}.}
    \label{fig:cumulative}
    \vspace{-10pt}
\end{figure}

\begin{figure}[t!]
    \centering
    \includegraphics[width=0.99\linewidth]{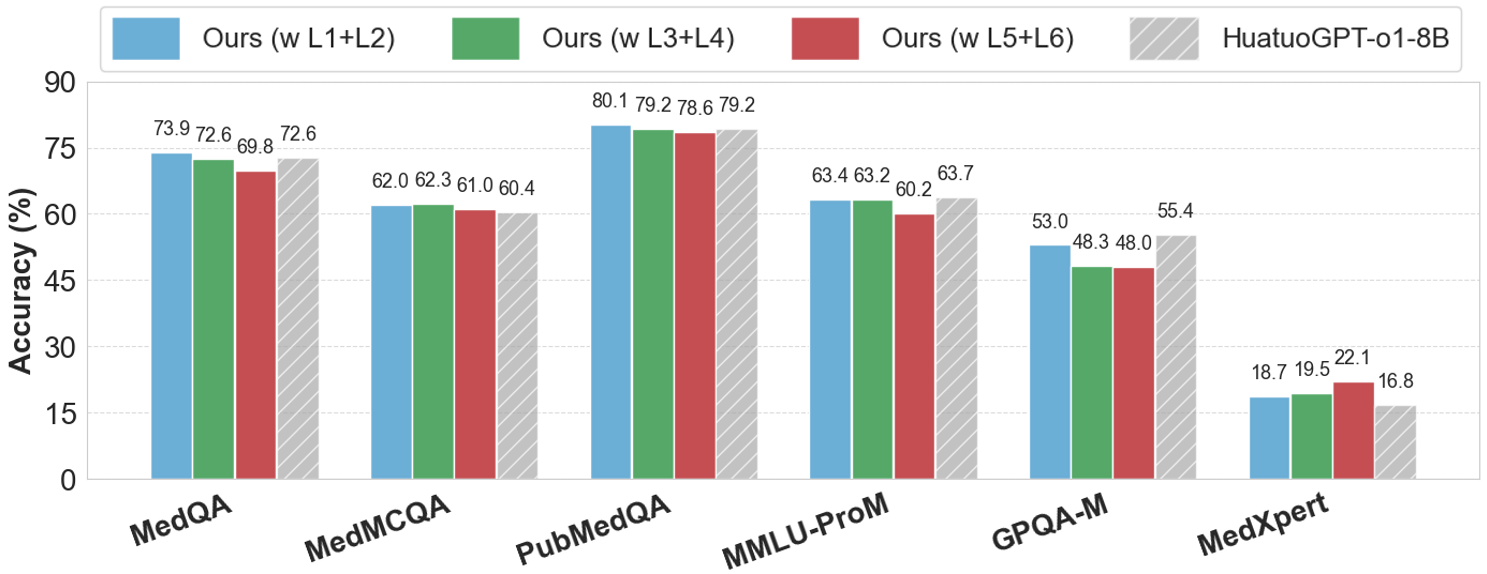}
    \vspace{-5pt}
    \caption{Performance on six benchmarks when training with distinct difficulty groups: easy (L1+L2), medium (L3+L4), and hard (L5+L6). While harder training data improves MedXpert~\cite{zuo2025medxpertqa} accuracy, performance on other benchmarks declines, suggesting that relying solely on difficult samples may impair general reasoning ability.}
    \label{fig:difficulty}
    \vspace{-17pt}
\end{figure}

\subsection{RQ3: Impact of Dataset Quality}
% \paragraph{Effect of Training Difficulty on Benchmark Performance.}
We analyze how increasing training difficulty affects performance across six benchmarks, as shown in Fig.~\ref{fig:cumulative}. MedQA, MedMCQA, and PubMedQA~\cite{medqa,medmcqa,jin2019pubmedqa} exhibit inverse U-shaped trends, 
performance peaks with moderate difficulty (L1–L4) and declines with harder samples (L5–L6),
suggesting diminishing returns from high-difficulty data. 
In contrast, MMLU-ProM~\cite{wang2024mmlupro} and GPQA-M~\cite{rein2024gpqa} show oscillating patterns, while MedXpert~\cite{zuo2025medxpertqa} improves steadily with increasing difficulty, highlighting the value of harder samples for complex tasks. To validate this, we train models on three difficulty groups (easy: L1+L2, medium: L3+L4, hard: L5+L6; Fig.~\ref{fig:difficulty}). On MedXpert~\cite{zuo2025medxpertqa}, models trained on hard samples perform best, confirming their role in promoting advanced reasoning. For other benchmarks, training on easy and medium levels yields better generalization, while hard-only training underperforms.

\paragraph{Emerging Reasoning Capability from Simple Data, Indicating Benchmark Limits.}
Interestingly, models trained only on L1+L2 (a total of 2,400 samples) already match or surpass HuatuoGPT-o1-8B~\cite{chen2024huatuogpt} on several benchmarks. As shown in Fig.~\ref{fig:cumulative}, even on MedXpert, only training with L1 data exceeds HuatuoGPT-o1-8B~\cite{chen2024huatuogpt}, with further gains from adding more levels, indicating that reasoning can emerge from simple data. These findings underscore the importance of balanced training difficulty to support broad generalization. They also reveal a potential pitfall: if high benchmark scores can be achieved without exposure to difficult samples, such scores may not reflect genuine reasoning ability, raising concerns about the adequacy of current benchmark designs.

\vspace{-5pt}
\begin{tcolorbox}[
    colback=answerbox,      % background color
    colframe=teal!60!black, % border color
    arc=6pt,    % rounded corners, like your example
    boxrule=0.8pt,          % border thickness
    left=2mm, right=2mm,    % padding
    top=1mm, bottom=1mm,
]
\textbf{Finding 3.1:}  
Mixed difficulty training is crucial for generalizable reasoning.  
\vspace{1mm}\\
\textbf{Finding 3.2:}  
Current benchmarks may insufficient to capture true reasoning progress.
\end{tcolorbox}
\vspace{-10pt}

\subsection{Main Results}

\begin{figure}[t!]
    \centering
    \begin{minipage}[t]{0.47\linewidth}
        \includegraphics[width=\linewidth]{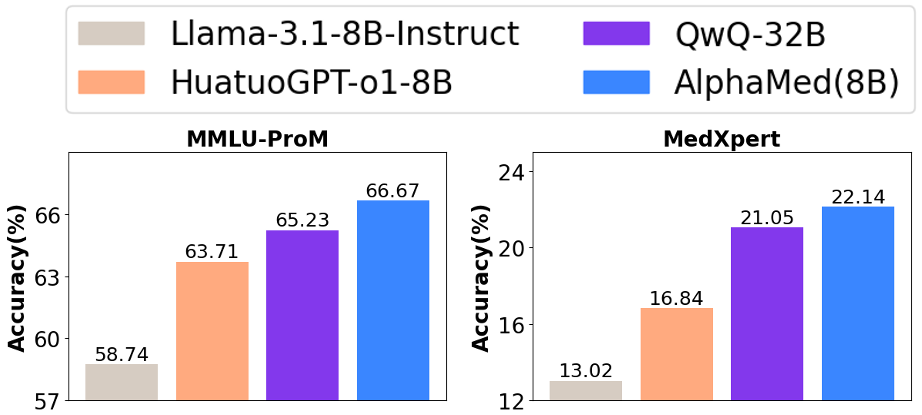}
        \caption{Comparison of AlphaMed(8B) with prior models on MMLU-ProM~\cite{wang2024mmlu} and MedXpert~\cite{zuo2025medxpertqa}. Despite its smaller scale and use of minimalist RL, AlphaMed(8B) outperforms the larger model QwQ-32B~\cite{qwq-32b-preview} and other baselines. 
        }
        \label{fig:8b}
    \end{minipage}%
    \hspace{0.02\linewidth}
    % \\
    \begin{minipage}[t]{0.47\linewidth}
        \includegraphics[width=\linewidth]{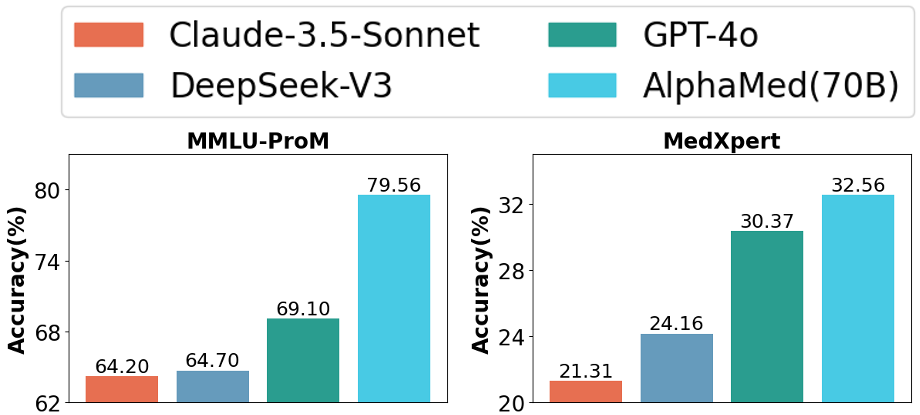}
        \caption{\textbf{AlphaMed(70B)} achieves superior performance over Claude-3.5-Sonnet~\cite{Claude3}, GPT-4o~\cite{gpt4o}, and DeepSeek-V3 (671B)~\cite{liu2024deepseek} on MMLU-ProM~\cite{wang2024mmlu} and MedXpert~\cite{zuo2025medxpertqa}, showcasing its strong reasoning ability.}
        \label{fig:70b}
    \end{minipage}
    \vspace{-20pt}
\end{figure}

Building on the above findings which highlight the importance of dataset quantity, diversity, informativeness, and mixed difficulty for incentivizing reasoning, we construct our final training set accordingly. Specifically, we include all samples from MedQA~\cite{medqa} due to its high informativeness, and sample 1,600 QA pairs from each difficulty level of MedMCQA~\cite{medmcqa} to match the overall scale of MedQA~\cite{medqa}. 
PubMedQA~\cite{jin2019pubmedqa} is excluded due to its limited informativeness and the performance degradation observed when it is included, as discussed in RQ1 and RQ2. The final training set comprises 19,178 QA pairs.
This dataset is used to train our final models: \textbf{AlphaMed(8B)}, based on \texttt{Llama3.1-8B-Instruct}, and \textbf{AlphaMed(70B)}, based on \texttt{Llama3.1-70B-Instruct}, both optimized using minimalist rule-based RL. Since MedQA~\cite{medqa} and MedMCQA~\cite{medmcqa} are used for training, we treat PubMedQA~\cite{jin2019pubmedqa}, MMLU-ProM~\cite{wang2024mmlupro}, GPQA-M~\cite{rein2023gpqa}, and MedXpert~\cite{zuo2025medxpertqa} as out-of-domain (OOD) benchmarks.

We present the full results in Tab.~\ref{tab:full_result}. Across both model scales, \texttt{AlphaMed} consistently outperforms all compared methods on both in-domain and OOD benchmarks, using only minimalist rule-based RL and multiple-choice QA supervision. Remarkably, this advantage holds even against models trained with more complex strategies~\cite{chen2024huatuogpt,zhang2024ultramedical}, including SFT on distilled CoT data~\cite{chen2024huatuogpt,zhang2024ultramedical,huang2025m1} and methods enhanced with test-time scaling~\cite{huang2025m1}.
Notably, \texttt{AlphaMed(8B)} surpasses the larger reasoning model \texttt{QwQ-32B}~\cite{qwq-32b-preview} on challenging OOD benchmarks, as shown in Fig.~\ref{fig:8b}. At the 70B scale, \texttt{AlphaMed(70B)} outperforms even closed-source models such as \texttt{GPT-4o}~\cite{gpt4o} and \texttt{Claude-3.5-Sonnet}~\cite{Claude3}, as well as the open-source \texttt{DeepSeek-V3} (671B parameters)~\cite{liu2024deepseek}, as shown in Fig.~\ref{fig:70b}.
These results show that minimalist rule-based RL, trained with a well-constructed multiple-choice QA dataset, enables effective and scalable medical reasoning in LLMs without relying on distilled CoT supervision.

\begin{table}[t!]
\centering
\scalebox{0.8}{
\begin{tabular}{lcccccc}
\toprule
\textbf{Model} & \textbf{MedQA} & \textbf{MedMCQA} & \textbf{PubMedQA} & \textbf{MMLU-ProM} & \textbf{GPQA-M} & \textbf{MedXpert} \\
\midrule
\multicolumn{1}{c}{} & \multicolumn{2}{c}{\textit{In-Domain}} & \multicolumn{4}{c}{\textit{Out-of-Domain}} \\
\cmidrule(lr){2-3} \cmidrule(lr){4-7}
\multicolumn{1}{l}{\textbf{\textit{Challenge Level}}} & \textit{Normal} & \textit{Normal} & \textit{Normal} & \textit{Hard} & \textit{Hard} & \textit{Hard+} \\
\midrule
\multicolumn{7}{c}{< 10B LLMs} \\
\midrule
Llama-3.1-8B-Instruct & 58.72 & 56.21 & 75.21 & 58.74 & 42.73 & 13.02 \\
Qwen2.5-7B-Instruct & 61.51 & 56.56 & 71.30 & 61.17 & 42.56 & 12.15 \\
Qwen2.5-7B-Instruct$^{+}$ & 64.49 & 56.11 & 72.60 & 62.15 & 52.56 & 13.18 \\
MedLlama3-8B-v1 & 55.07 & 34.74 & 52.70 & 27.43 & 30.77 & 11.04 \\
MedLlama3-8B-v2 & 59.39 & 59.34 & 75.50 & 55.11 & 36.41 & 13.46 \\
MMed-8B$^{\dag\ddag}$ & 54.28 & 52.71 & 63.40 & 48.27 & 34.87 & 13.73 \\
MMedS-8B$^{\dag\ddag}$ & 57.19 & 47.29 & 77.50 & 33.55 & 22.05 & 17.39 \\
MMed-8B-EnIns$^{\dag\ddag}$ & 60.33 & 58.09 & 63.80 & 51.60 & 45.90 & 18.56 \\
Med42-8B$^{\ddag}$ & 59.78 & 56.35 & 76.00 & 55.64 & 48.21 & 14.63 \\
OpenBioLLM-8B$^{\dag\ddag\lozenge}$ & 55.30 & 54.63 & 70.10 & 49.32 & 41.03 & 14.29 \\
UltraMedical-8B-3$^{\dag\ddag\lozenge}$ & 71.09 & 59.22 & 71.00 & 61.50 & 50.00 & 15.25 \\
UltraMedical-8B-3.1$^{\dag\ddag\lozenge}$ & 75.73 & 63.78 & 79.20 & 64.30 & 48.72 & 17.39 \\
HuatuoGPT-o1-8B$^{\dag\ddag\lozenge}$ & 72.60 & 60.40 & 79.20 & 63.71 & 55.38 & 16.84 \\
\textcolor{darkgray}{m1-7B$^{\dag\ddag}$} & \textcolor{darkgray}{75.81} & \textcolor{darkgray}{62.54} & \textcolor{darkgray}{75.80} & \textcolor{darkgray}{65.86} & \textcolor{darkgray}{53.08} & \textcolor{darkgray}{19.81} \\
\textbf{AlphaMed(8B)} & \textbf{76.19} & \textbf{64.47} & \textbf{80.40} & \textbf{66.67} & \textbf{58.44} & \textbf{22.14} \\
\midrule
\multicolumn{7}{c}{> 10B LLMs} \\
\midrule
Llama-3.1-70B-Instruct & 78.42 & 72.53 & 78.52 & 74.50 & 55.73 & 21.32 \\
QwQ-32B & 78.62 & 69.71 & 77.85 & 65.23 & 56.92 & 21.05 \\
Qwen2.5-32B-Instruct & 75.26 & 64.83 & 68.00 & 74.72 & 63.85 & 13.87 \\
Qwen2.5-32B-Instruct$^{+}$ & 74.86 & 64.33 & 68.90 & 74.72 & 64.87 & 14.56 \\
Qwen2.5-72B-Instruct & 74.55 & 66.60 & 70.80 & 66.06 & 62.05 & 14.91 \\
Qwen2.5-72B-Instruct$^{+}$ & 76.43 & 66.15 & 71.30 & 69.77 & 63.85 & 19.65 \\
Med42-70B$^{\ddag}$ & 51.14 & 62.28 & 78.10 & 54.53 & 50.77 & 16.29 \\
OpenBioLLM-70B$^{\dag\ddag\lozenge}$ & 75.10 & 74.23 & 79.30 & 71.92 & 50.77 & 21.33 \\
UltraMedical-70B-3$^{\dag\ddag\lozenge}$ & 83.90 & 72.94 & 80.00 & 73.94 & 58.72 & 21.67 \\
HuatuoGPT-o1-70B$^{\dag\ddag\lozenge}$ & 83.30 & 73.60 & 80.60 & 76.09 & 66.67 & 26.36 \\
\textcolor{darkgray}{m1-32B$^{\dag\ddag}$} & \textcolor{darkgray}{83.50} & \textcolor{darkgray}{67.34} & \textcolor{darkgray}{77.60} & \textcolor{darkgray}{77.94} & \textcolor{darkgray}{66.67} & \textcolor{darkgray}{25.53} \\
\textbf{AlphaMed(70B)} & \textbf{87.52} & \textbf{75.09} & \textbf{80.90} & \textbf{79.56} & \textbf{77.46} & \textbf{32.56} \\
\bottomrule
\end{tabular}
}
\vspace{2mm}
\caption{Combined performance of models on six medical QA benchmarks with varying levels of challenge. In-domain and out-of-domain tasks, as well as challenge levels (Normal, Hard, Hard+), are indicated below the task names. 
\textcolor{darkgray}{\textbf{m1}} denotes models that use \textit{test-time scaling during inference}. 
$^{+}$: using CoT prompting during inference; 
$^{\dag}$: trained with distilled CoT data from stronger models (e.g., GPT-4o); 
$^{\ddag}$: trained with external datasets beyond MedQA and MedMCQA; 
$^{\lozenge}$: trained via RL with verifier reward models or distilled preference data from powerful models (e.g., GPT-4o). 
\textbf{AlphaMed (Ours)} is trained solely with minimalist rule-based RL on multi-choice QA, without any SFT on distilled CoT data, preference data, or rewards from verifiers.
}
\label{tab:full_result}
\vspace{-7mm}
\end{table}

\section{Conclusion}
We present AlphaMed, the first work to demonstrate that reasoning capabilities can emerge solely through minimalist rule-based RL, without relying on SFT with distilled CoT data. By leveraging only multiple-choice QA datasets, AlphaMed achieves state-of-the-art performance across six diverse and challenging medical QA benchmarks, surpassing models trained with conventional SFT+RL pipelines, and even outperforming closed-source models (e.g., GPT-4o~\cite{gpt4o}. Through comprehensive data-centric analyses, we show that reasoning ability can be effectively incentivized by selecting data based on informativeness. We further find that increasing the number of informative training samples improves performance, and that varying difficulty levels contribute differently across benchmarks, underscoring the importance of mixing difficulty to promote generalizable reasoning. A well-curated dataset with high informativeness and diverse difficulty levels is key to advancing reasoning, without requiring handcrafted rationales or distilled data from closed models. Our findings also reveal a critical caveat: while challenging benchmarks benefit from harder training samples, others exhibit mixed or plateauing trends, suggesting that existing benchmarks may be insufficient to evaluate progress of reasoning ability. This highlights the need for more challenging, reasoning-oriented benchmarks. Altogether, AlphaMed not only establishes a strong medical LLM, but also offers insights into how models reach final predictions through emergent reasoning, encouraging further exploration of interpretable systems in medical NLP.

\clearpage
{\small
\bibliographystyle{IEEEtran}
\bibliography{ref.bib}

% Generated by IEEEtran.bst, version: 1.14 (2015/08/26)
\begin{thebibliography}{10}
\providecommand{\url}[1]{#1}
\csname url@samestyle\endcsname
\providecommand{\newblock}{\relax}
\providecommand{\bibinfo}[2]{#2}
\providecommand{\BIBentrySTDinterwordspacing}{\spaceskip=0pt\relax}
\providecommand{\BIBentryALTinterwordstretchfactor}{4}
\providecommand{\BIBentryALTinterwordspacing}{\spaceskip=\fontdimen2\font plus
\BIBentryALTinterwordstretchfactor\fontdimen3\font minus \fontdimen4\font\relax}
\providecommand{\BIBforeignlanguage}[2]{{%
\expandafter\ifx\csname l@#1\endcsname\relax
\typeout{** WARNING: IEEEtran.bst: No hyphenation pattern has been}%
\typeout{** loaded for the language `#1'. Using the pattern for}%
\typeout{** the default language instead.}%
\else
\language=\csname l@#1\endcsname
\fi
#2}}
\providecommand{\BIBdecl}{\relax}
\BIBdecl

\bibitem{o1journey}
Y.~Qin, X.~Li, H.~Zou, Y.~Liu, S.~Xia, Z.~Huang, Y.~Ye, W.~Yuan, H.~Liu, Y.~Li \emph{et~al.}, ``O1 replication journey: A strategic progress report--part 1,'' \emph{arXiv preprint arXiv:2410.18982}, 2024.

\bibitem{o1p1}
Z.~Zeng, Q.~Cheng, Z.~Yin, B.~Wang, S.~Li, Y.~Zhou, Q.~Guo, X.~Huang, and X.~Qiu, ``Scaling of search and learning: A roadmap to reproduce o1 from reinforcement learning perspective,'' \emph{arXiv preprint arXiv:2412.14135}, 2024.

\bibitem{o1p9}
J.~Wang, M.~Fang, Z.~Wan, M.~Wen, J.~Zhu, A.~Liu, Z.~Gong, Y.~Song, L.~Chen, L.~M. Ni \emph{et~al.}, ``Openr: An open source framework for advanced reasoning with large language models,'' \emph{arXiv preprint arXiv:2410.09671}, 2024.

\bibitem{guan2024deliberative}
\BIBentryALTinterwordspacing
M.~Y. Guan, M.~Joglekar, E.~Wallace, S.~Jain, B.~Barak, A.~Heylar, R.~Dias, A.~Vallone, H.~Ren, J.~Wei, H.~W. Chung, S.~Toyer, J.~Heidecke, A.~Beutel, and A.~Glaese, ``Deliberative alignment: Reasoning enables safer language models,'' \emph{OpenAI Blog}, 2024. [Online]. Available: \url{https://openai.com/index/deliberative-alignment/}
\BIBentrySTDinterwordspacing

\bibitem{saab2024capabilities}
K.~Saab, T.~Tu, W.-H. Weng, R.~Tanno, D.~Stutz, E.~Wulczyn, F.~Zhang, T.~Strother, C.~Park, E.~Vedadi \emph{et~al.}, ``Capabilities of gemini models in medicine,'' \emph{arXiv preprint arXiv:2404.18416}, 2024.

\bibitem{chen2024cod}
J.~Chen, C.~Gui, A.~Gao, K.~Ji, X.~Wang, X.~Wan, and B.~Wang, ``Cod, towards an interpretable medical agent using chain of diagnosis,'' \emph{arXiv preprint arXiv:2407.13301}, 2024.

\bibitem{patel2005thinking}
V.~L. Patel, J.~F. Arocha, and J.~Zhang, ``Thinking and reasoning in medicine,'' \emph{The Cambridge handbook of thinking and reasoning}, vol.~14, pp. 727--750, 2005.

\bibitem{o12}
S.~Xu, Y.~Zhou, Z.~Liu, Z.~Wu, T.~Zhong, H.~Zhao, Y.~Li, H.~Jiang, Y.~Pan, J.~Chen \emph{et~al.}, ``Towards next-generation medical agent: How o1 is reshaping decision-making in medical scenarios,'' \emph{arXiv preprint arXiv:2411.14461}, 2024.

\bibitem{o1p10}
M.-H. Temsah, A.~Jamal, K.~Alhasan, A.~A. Temsah, and K.~H. Malki, ``Openai o1-preview vs. chatgpt in healthcare: A new frontier in medical ai reasoning,'' \emph{Cureus}, vol.~16, no.~10, p. e70640, 2024.

\bibitem{o1medicalpreliminary}
Y.~Xie, J.~Wu, H.~Tu, S.~Yang, B.~Zhao, Y.~Zong, Q.~Jin, C.~Xie, and Y.~Zhou, ``A preliminary study of o1 in medicine: Are we closer to an ai doctor?'' \emph{arXiv preprint arXiv:2409.15277}, 2024.

\bibitem{chen2023huatuogpt}
J.~Chen, X.~Wang, K.~Ji, A.~Gao, F.~Jiang, S.~Chen, H.~Zhang, D.~Song, W.~Xie, C.~Kong \emph{et~al.}, ``Huatuogpt-ii, one-stage training for medical adaption of llms,'' \emph{arXiv preprint arXiv:2311.09774}, 2023.

\bibitem{wei2022chainofthought}
J.~Wei, X.~Wang, D.~Schuurmans, M.~Bosma, B.~Ichter, F.~Xia, E.~H. Chi, Q.~V. Le, and D.~Zhou, ``Chain-of-thought prompting elicits reasoning in large language models,'' \emph{Advances in Neural Information Processing Systems}, vol.~35, pp. 24\,824--24\,837, 2022.

\bibitem{chung2022scaling}
H.~W. Chung, L.~Hou, S.~Longpre, B.~Zoph, Y.~Tay, W.~Fedus, Y.~Li, X.~Wang, J.~Wei \emph{et~al.}, ``Scaling instruction-finetuned language models,'' \emph{arXiv preprint arXiv:2210.11416}, 2022.

\bibitem{zelikman2022star}
E.~Zelikman, Y.~Wu, J.~Mu, and N.~D. Goodman, ``Star: Bootstrapping reasoning with reasoning,'' \emph{Advances in Neural Information Processing Systems}, vol.~35, pp. 15\,476--15\,488, 2022.

\bibitem{chu2025sft}
T.~Chu, Y.~Zhai, J.~Yang, S.~Tong, S.~Xie, D.~Schuurmans, Q.~V. Le, S.~Levine, and Y.~Ma, ``Sft memorizes, rl generalizes: A comparative study of foundation model post-training,'' \emph{arXiv preprint arXiv:2501.17161}, 2025.

\bibitem{vl-rethinker}
H.~Wang, C.~Qu, Z.~Huang, W.~Chu, F.~Lin, and W.~Chen, ``Vl-rethinker: Incentivizing self-reflection of vision-language models with reinforcement learning,'' \emph{arXiv preprint arXiv:2504.08837}, 2025.

\bibitem{ouyang2022training}
L.~Ouyang, J.~Wu, X.~Jiang, D.~Almeida, C.~Wainwright, P.~Mishkin, C.~Zhang, S.~Agarwal, K.~Slama, A.~Ray \emph{et~al.}, ``Training language models to follow instructions with human feedback,'' \emph{Advances in Neural Information Processing Systems}, vol.~35, pp. 27\,730--27\,744, 2022.

\bibitem{wang2025learning}
H.~Wang, L.~Li, C.~Qu, F.~Zhu, W.~Xu, W.~Chu, and F.~Lin, ``Learning autonomous code integration for math language models,'' \emph{arXiv preprint arXiv:2502.00691}, 2025.

\bibitem{rafailov2023direct}
R.~Rafailov, A.~Sharma, E.~Mitchell, C.~D. Manning, S.~Ermon, and C.~Finn, ``Direct preference optimization: Your language model is secretly a reward model,'' \emph{Advances in Neural Information Processing Systems}, vol.~36, 2023.

\bibitem{ura2024openbiollm}
A.~Ura, ``Openbiollm-70b: Advancing open-source biomedical llms with direct preference optimization,'' \emph{Hugging Face Blog}, 2024, available at \url{https://huggingface.co/blog/aaditya/openbiollm}.

\bibitem{guo2025deepseek}
D.~Guo, D.~Yang, H.~Zhang, J.~Song, R.~Zhang, R.~Xu, Q.~Zhu, S.~Ma, P.~Wang, X.~Bi \emph{et~al.}, ``Deepseek-r1: Incentivizing reasoning capability in llms via reinforcement learning,'' \emph{arXiv preprint arXiv:2501.12948}, 2025.

\bibitem{zeng2025acecoder}
H.~Zeng, D.~Jiang, H.~Wang, P.~Nie, X.~Chen, and W.~Chen, ``Acecoder: Acing coder rl via automated test-case synthesis,'' \emph{arXiv preprint arXiv:2502.01718}, 2025.

\bibitem{pan2025medvlm}
J.~Pan, C.~Liu, J.~Wu, F.~Liu, J.~Zhu, H.~B. Li, C.~Chen, C.~Ouyang, and D.~Rueckert, ``Medvlm-r1: Incentivizing medical reasoning capability of vision-language models (vlms) via reinforcement learning,'' \emph{arXiv preprint arXiv:2502.19634}, 2025.

\bibitem{zhang2023huatuogpt}
H.~Zhang, J.~Chen, F.~Jiang, F.~Yu, Z.~Chen, J.~Li, G.~Chen, X.~Wu, Z.~Zhang, Q.~Xiao \emph{et~al.}, ``Huatuogpt, towards taming language model to be a doctor,'' \emph{arXiv preprint arXiv:2305.15075}, 2023.

\bibitem{labrak2024biomistral}
Y.~Labrak, A.~Bazoge, E.~Morin, P.-A. Gourraud, M.~Rouvier, and R.~Dufour, ``Biomistral: A collection of open-source pretrained large language models for medical domains,'' \emph{arXiv preprint arXiv:2402.10373}, 2024.

\bibitem{ke2023continual}
Z.~Ke, Y.~Shao, H.~Lin, T.~Konishi, G.~Kim, and B.~Liu, ``Continual pre-training of language models,'' \emph{arXiv preprint arXiv:2302.03241}, 2023.

\bibitem{zhang2024ultramedical}
K.~Zhang, S.~Zeng, E.~Hua, N.~Ding, Z.-R. Chen, Z.~Ma, H.~Li, G.~Cui, B.~Qi, X.~Zhu \emph{et~al.}, ``Ultramedical: Building specialized generalists in biomedicine,'' \emph{Advances in Neural Information Processing Systems}, vol.~37, pp. 26\,045--26\,081, 2024.

\bibitem{chen2024huatuogpt-o1}
J.~Chen, Z.~Cai, K.~Ji, X.~Wang, W.~Liu, R.~Wang, J.~Hou, and B.~Wang, ``Huatuogpt-o1: Towards medical complex reasoning with llms,'' \emph{arXiv preprint arXiv:2412.18925}, 2024.

\bibitem{huang2025m1}
X.~Huang, J.~Wu, H.~Liu, X.~Tang, and Y.~Zhou, ``m1: Unleash the potential of test-time scaling for medical reasoning with large language models,'' \emph{arXiv preprint arXiv:2504.00869}, 2025.

\bibitem{ppo2017}
J.~Schulman, F.~Wolski, P.~Dhariwal, A.~Radford, and O.~Klimov, ``Proximal policy optimization algorithms,'' \emph{arXiv preprint arXiv:1707.06347}, 2017.

\bibitem{sheng2024hybridflow}
G.~Sheng, C.~Zhang, Z.~Ye, X.~Wu, W.~Zhang, R.~Zhang, Y.~Peng, H.~Lin, and C.~Wu, ``Hybridflow: A flexible and efficient rlhf framework,'' \emph{arXiv preprint arXiv:2409.19256}, 2024.

\bibitem{jin2021disease}
D.~Jin, E.~Pan, N.~Oufattole, W.-H. Weng, H.~Fang, and P.~Szolovits, ``What disease does this patient have? a large-scale open domain question answering dataset from medical exams,'' \emph{Applied Sciences}, vol.~11, no.~14, p. 6421, 2021.

\bibitem{pal2022medmcqa}
A.~Pal, L.~K. Umapathi, and M.~Sankarasubbu, ``Medmcqa: A large-scale multi-subject multi-choice dataset for medical domain question answering,'' in \emph{Conference on health, inference, and learning}.\hskip 1em plus 0.5em minus 0.4em\relax PMLR, 2022, pp. 248--260.

\bibitem{jin2019pubmedqa}
Q.~Jin, B.~Dhingra, Z.~Liu, W.~W. Cohen, and X.~Lu, ``Pubmedqa: A dataset for biomedical research question answering,'' \emph{arXiv preprint arXiv:1909.06146}, 2019.

\bibitem{wang2024mmlu}
Y.~Wang, X.~Ma, G.~Zhang, Y.~Ni, A.~Chandra, S.~Guo, W.~Ren, A.~Arulraj, X.~He, Z.~Jiang \emph{et~al.}, ``Mmlu-pro: A more robust and challenging multi-task language understanding benchmark,'' in \emph{The Thirty-eight Conference on Neural Information Processing Systems Datasets and Benchmarks Track}, 2024.

\bibitem{rein2024gpqa}
D.~Rein, B.~L. Hou, A.~C. Stickland, J.~Petty, R.~Y. Pang, J.~Dirani, J.~Michael, and S.~R. Bowman, ``Gpqa: A graduate-level google-proof q\&a benchmark,'' in \emph{First Conference on Language Modeling}, 2024.

\bibitem{zuo2025medxpertqa}
Y.~Zuo, S.~Qu, Y.~Li, Z.~Chen, X.~Zhu, E.~Hua, K.~Zhang, N.~Ding, and B.~Zhou, ``Medxpertqa: Benchmarking expert-level medical reasoning and understanding,'' \emph{arXiv preprint arXiv:2501.18362}, 2025.

\bibitem{tang2025medagentsbench}
X.~Tang, D.~Shao, J.~Sohn, J.~Chen, J.~Zhang, J.~Xiang, F.~Wu, Y.~Zhao, C.~Wu, W.~Shi \emph{et~al.}, ``Medagentsbench: Benchmarking thinking models and agent frameworks for complex medical reasoning,'' \emph{arXiv preprint arXiv:2503.07459}, 2025.

\bibitem{wang2024mmlupro}
Y.~Wang, X.~Ma, G.~Zhang, Y.~Ni, A.~Chandra, S.~Guo, W.~Ren, A.~Arulraj, X.~He, Z.~Jiang \emph{et~al.}, ``Mmlu-pro: A more robust and challenging multi-task language understanding benchmark,'' \emph{arXiv preprint arXiv:2406.01574}, 2024.

\bibitem{pal2024openbiollms}
M.~S.~A. Pal and M.~Sankarasubbu, ``Openbiollms: Advancing open-source large language models for healthcare and life sciences,'' 2024.

\bibitem{qiu2024towards}
P.~Qiu, C.~Wu, X.~Zhang, W.~Lin, H.~Wang, Y.~Zhang, Y.~Wang, and W.~Xie, ``Towards building multilingual language model for medicine,'' \emph{Nature Communications}, vol.~15, no.~1, p. 8384, 2024.

\bibitem{christophe2024med42}
C.~Christophe, P.~K. Kanithi, P.~Munjal, T.~Raha, N.~Hayat, R.~Rajan, A.~Al-Mahrooqi, A.~Gupta, M.~U. Salman, G.~Gosal \emph{et~al.}, ``Med42--evaluating fine-tuning strategies for medical llms: Full-parameter vs. parameter-efficient approaches,'' \emph{arXiv preprint arXiv:2404.14779}, 2024.

\bibitem{medqa}
D.~Jin, E.~Pan, N.~Oufattole, W.-H. Weng, H.~Fang, and P.~Szolovits, ``What disease does this patient have? a large-scale open domain question answering dataset from medical exams,'' \emph{Applied Sciences}, vol.~11, no.~14, p. 6421, 2021.

\bibitem{medmcqa}
A.~Pal, L.~K. Umapathi, and M.~Sankarasubbu, ``Medmcqa: A large-scale multi-subject multi-choice dataset for medical domain question answering,'' in \emph{Conference on Health, Inference, and Learning}.\hskip 1em plus 0.5em minus 0.4em\relax PMLR, 2022, pp. 248--260.

\bibitem{llama3}
A.~Dubey, A.~Jauhri, A.~Pandey, A.~Kadian, A.~Al-Dahle, A.~Letman, A.~Mathur, A.~Schelten, A.~Yang, A.~Fan \emph{et~al.}, ``The llama 3 herd of models,'' \emph{arXiv preprint arXiv:2407.21783}, 2024.

\bibitem{chen2024huatuogpt}
J.~Chen, Z.~Cai, K.~Ji, X.~Wang, W.~Liu, R.~Wang, J.~Hou, and B.~Wang, ``Huatuogpt-o1, towards medical complex reasoning with llms,'' \emph{arXiv preprint arXiv:2412.18925}, 2024.

\bibitem{gpt4o}
A.~Hurst, A.~Lerer, A.~P. Goucher, A.~Perelman, A.~Ramesh, A.~Clark, A.~Ostrow, A.~Welihinda, A.~Hayes, A.~Radford \emph{et~al.}, ``Gpt-4o system card,'' \emph{arXiv preprint arXiv:2410.21276}, 2024.

\bibitem{qwq-32b-preview}
\BIBentryALTinterwordspacing
Q.~Team, ``Qwq: Reflect deeply on the boundaries of the unknown,'' November 2024. [Online]. Available: \url{https://qwenlm.github.io/blog/qwq-32b-preview/}
\BIBentrySTDinterwordspacing

\bibitem{Claude3}
Anthropic, ``The claude 3 model family: Opus, sonnet, haiku,'' \url{https://www-cdn.anthropic.com/de8ba9b01c9ab7cbabf5c33b80b7bbc618857627/Model_Card_Claude_3.pdf}, 2024.

\bibitem{liu2024deepseek}
A.~Liu, B.~Feng, B.~Xue, B.~Wang, B.~Wu, C.~Lu, C.~Zhao, C.~Deng, C.~Zhang, C.~Ruan \emph{et~al.}, ``Deepseek-v3 technical report,'' \emph{arXiv preprint arXiv:2412.19437}, 2024.

\bibitem{rein2023gpqa}
D.~Rein, B.~L. Hou, A.~C. Stickland, J.~Petty, R.~Y. Pang, J.~Dirani, J.~Michael, and S.~R. Bowman, ``Gpqa: A graduate-level google-proof q\&a benchmark,'' \emph{arXiv preprint arXiv:2311.12022}, 2023.

\bibitem{qwen2.5}
\BIBentryALTinterwordspacing
Q.~Team, ``Qwen2.5: A party of foundation models,'' September 2024. [Online]. Available: \url{https://qwenlm.github.io/blog/qwen2.5/}
\BIBentrySTDinterwordspacing

\end{thebibliography}
}

%%%%%%%%%%%%%%%%%%%%%%%%%%%%%%%%%%%%%%%%%%%%%%%%%%%%%%%%%%%%
\clearpage
\appendix 
\section*{Limitations and Future Work}
Although AlphaMed achieves impressive results on multiple-choice QA tasks, its capabilities remain constrained by the closed-form nature of these benchmarks. Our evaluations are primarily conducted on existing mainstream medical QA datasets, all of which are close-ended and may not fully capture the spectrum of real-world clinical reasoning. Due to limitations in the current research landscape, it is challenging to systematically assess our model’s performance on open-ended QA tasks, which not only lack well-established benchmarks but are also inherently subjective, often requiring human evaluation for meaningful assessment. In future work, we aim to design and release open-ended benchmarks that involve human-in-the-loop evaluation, enabling more comprehensive and nuanced assessments of reasoning and decision-making in medical LLMs.

\section*{Broader Impact}
This work demonstrates that the reasoning capability of medical LLMs can be effectively incentivized using only multiple-choice QA data with minimalist rule-based RL, removing the need for SFT on costly distilled CoT data. By eliminating reliance on manual annotation and closed-source supervision, our approach substantially reduces the human effort and resources required for developing high-performing clinical models. However, the emerging reasoning processes in LLMs are inherently difficult to evaluate, as there is often no single “ground truth” reasoning path—especially in medicine, where multiple valid clinical justifications may exist for a single decision. Nonetheless, exposing these intermediate reasoning steps provides an important opportunity to observe and audit model behavior, ultimately encouraging the development of more transparent and trustworthy medical LLMs.

\section{Appendix}
\subsection{Difficulty Level Distribution}
\label{sec:diff dist}
To explore how the difficulty level of training data affects model performance, we annotate each sample by its response consistency across five inference passes of \texttt{Llama3.1-8B-Instruct}~\cite{llama3}. Specifically, L1 denotes samples where the model answers all attempts correctly (easy), while L6 includes those where all predictions are incorrect (hard). Intermediate levels (L2–L5) indicate varying degrees of partial correctness. Tab.~\ref{sec:diff dist} summarizes the distribution across MedQA\footnote{https://huggingface.co/datasets/GBaker/MedQA-USMLE-4-options-hf}, MedMCQA\footnote{https://huggingface.co/datasets/openlifescienceai/medmcqa}, and PubMedQA\footnote{https://huggingface.co/datasets/qiaojin/PubMedQA}.

\begin{table}[ht!]
\centering
\caption{Difficulty Level Distribution. L1 indicates samples where \texttt{Llama3.1-8B-Instruct}~\cite{llama3} predicts correctly in all 5 inference attempts (easiest), while L6 corresponds to samples where all predictions are incorrect (hardest). Intermediate levels (L2–L5) reflect partial correctness across attempts.}
\scalebox{0.99}{
\begin{tabular}{lccccccc}
\toprule
\textbf{Dataset} & \textbf{Total} & \textbf{L1} & \textbf{L2} & \textbf{L3} & \textbf{L4} & \textbf{L5} & \textbf{L6} \\
\midrule
MedQA      & 10,178   & 1,970  & 1,471  &   934  &   697  &   713  & 4,393 \\
MedMCQA    & 182,822  & 63,292 & 25,736 & 14,498 &  9,922 & 10,088 & 59,286 \\
PubMedQA   & 211,268  & 97,790 & 41,604 & 18,596 & 10,759 &  9,217 & 33,303 \\
\bottomrule
\end{tabular}
}
\label{tab:diff dist}
\end{table}

\subsection{Details of Evaluation Datasets}
\label{sec:eval data}
To thoroughly assess performance across varying levels of challenge, we evaluate on six medical QA benchmarks, grouped by challenge level:

\textbf{Normal challenge level}
\begin{itemize}
  \item MedQA~\cite{medqa}: A benchmark derived from US medical licensing exam questions, assessing clinical knowledge across a wide range of topics. Evaluation is based on the standard test split.

    \item MedMCQA~\cite{medmcqa}: A medical QA dataset based on entrance exams, designed to test foundational medical knowledge through multiple-choice questions. The official test split is used.
    
    \item PubMedQA~\cite{jin2019pubmedqa}: A biomedical question answering dataset where models choose from three fixed options, yes, no, or maybe, based on associated research abstracts, emphasizing factual understanding in biomedical literature. The official test split is used.
\end{itemize}

\textbf{Hard challenge level}
\begin{itemize}
  \item MMLU-ProM~\cite{wang2024mmlupro}: MMLU-ProM is the medical category subset of a broad multitask benchmark, focusing on professional-level medicine and related domains. Evaluation is conducted using the standard split established in~\cite{chen2024huatuogpt}.

  \item GPQA-M~\cite{rein2024gpqa}: It represents the biomedical subset of a graduate-level QA benchmark, featuring expert-curated questions intentionally designed to resist superficial retrieval and demand deep analytical reasoning. The evaluation follows the split from~\cite{chen2024huatuogpt}.
\end{itemize}

\textbf{Hard+ challenge level}
\begin{itemize}
    \item MedXpert~\cite{zuo2025medxpertqa}: A challenging benchmark designed to assess expert-level medical knowledge, clinical understanding, and complex reasoning. It covers diverse specialties and body systems, incorporates board-style exam questions, and is curated through expert review to ensure high difficulty, accuracy, and relevance to real-world medical decision-making.
\end{itemize}

\subsection{Effect of LLM Backbones}
\label{sec: ablate llm}
\begin{figure}[ht!]
    \centering
    \includegraphics[width=0.99\linewidth]{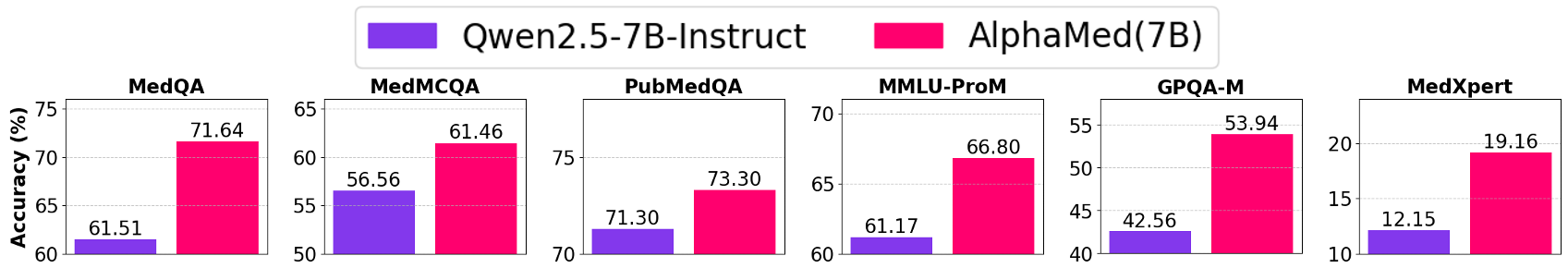}
    \caption{Performance comparison across six medical QA benchmarks. AlphaMed(7B) is initialized from \texttt{Qwen2.5-7B-Instruct}~\cite{qwen2.5} and trained using our constructed training set and minimalist rule-based RL pipeline. It achieves consistent improvements over the base model on all benchmarks.}
    \label{fig:qwen}
\end{figure}

To assess the generality of our proposed training pipeline and data design, we further apply the same minimalist rule-based RL approach, originally used for \texttt{Llama3.1-8B-Instruct}, to \texttt{Qwen2.5-7B-Instruct}~\cite{qwen2.5}. After training, the resulting AlphaMed(7B) model achieves consistent improvements across all six benchmarks, as shown in Fig.~\ref{fig:qwen}. Notably, the gains are particularly substantial on the more challenging datasets, MMLU-ProM~\cite{wang2024mmlu}, GPQA-M~\cite{rein2024gpqa}, and MedXpert~\cite{zuo2025medxpertqa}, demonstrating the robustness of our training strategy in enhancing medical reasoning. These results demonstrate that minimalist rule-based RL can incentivize reasoning capabilities and boost performance, exhibiting robustness across different backbone models.

\subsection{Success on Small LLM}
\label{sec: small llm}
\begin{figure}[ht!]
    \centering
    \includegraphics[width=0.99\linewidth]{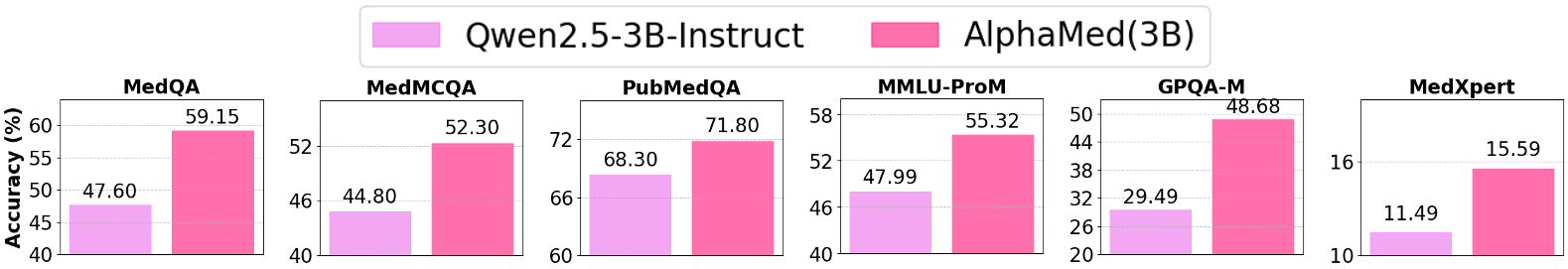}
            \caption{Performance comparison across six medical QA benchmarks. AlphaMed(3B) is initialized from \texttt{Qwen2.5-3B-Instruct}~\cite{qwen2.5} and trained with our constructed dataset using a minimalist rule-based RL pipeline. It achieves consistent gains over the base model.}
    \label{fig:3b}
\end{figure}

To further evaluate the effectiveness of our minimalist RL pipeline, we apply it to a small language model, \texttt{Qwen2.5-3B-Instruct}~\cite{qwen2.5}. As shown in Fig.~\ref{fig:3b}, our approach consistently improves performance across all six medical benchmarks, including substantial gains on \textbf{MedQA (+11.55\%)}, \textbf{GPQA-M (+19.19\%)}, and \textbf{MedXpert (+4.10\%)}. These results demonstrate that our RL framework can effectively incentivize reasoning capabilities even in smaller-scale models, and is not limited to large foundation models.

\subsection{Qualitative Results}
\label{sec: vis}
\begin{figure}[ht!]
    \centering
    \begin{minipage}{0.99\linewidth}
        \centering
        \includegraphics[width=\linewidth]{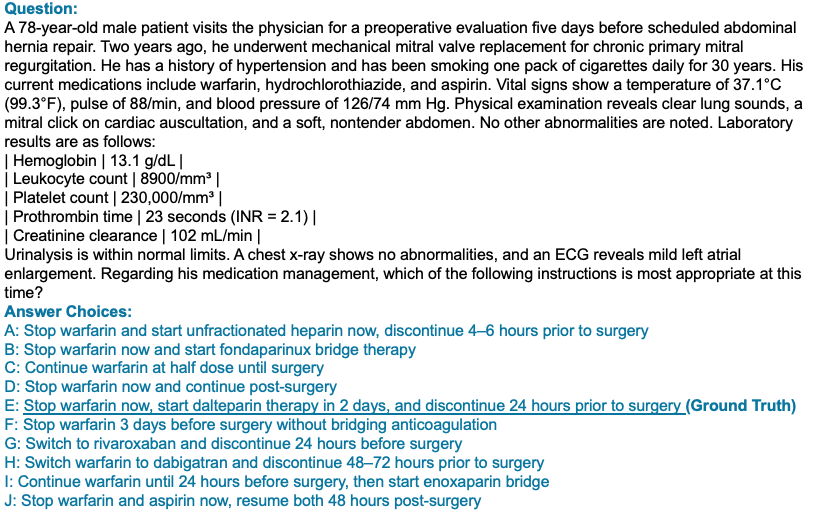}
    \end{minipage}
    \vspace{1ex} % vertical space between images
    \\
    \begin{minipage}{0.99\linewidth}
        \centering
        \includegraphics[width=\linewidth]{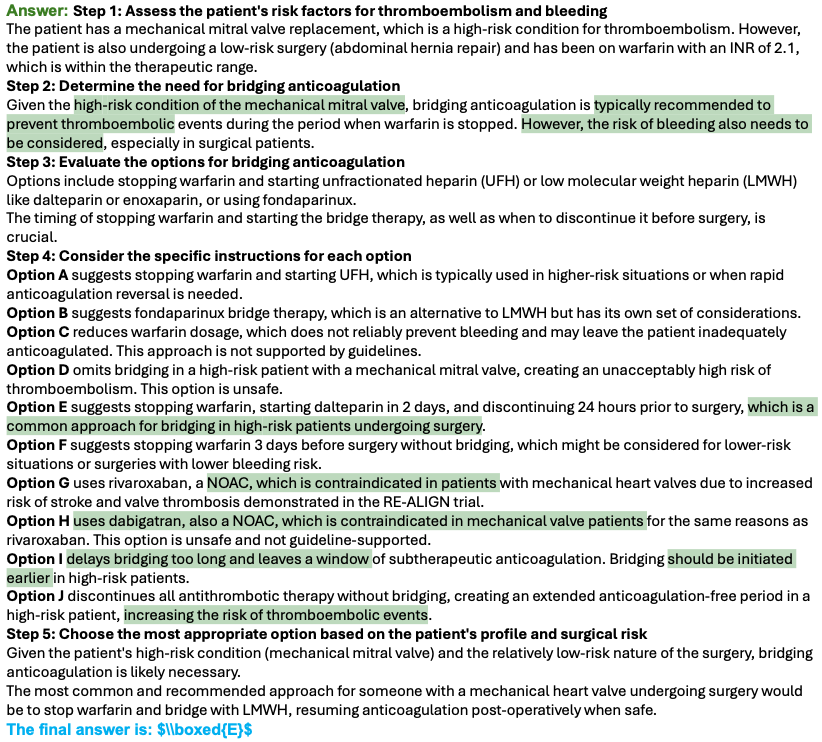}
    \end{minipage}
    \caption{Question and answer pair for Case 1. 
    \textcolor{cyan}{\textbf{Cyan text}} highlights the final predicted choices. 
    \colorbox{clinicalgreen}{\textcolor{black}{Green highlight}} are used to emphasize reasoning steps and key clinically key information.
    }
    \label{fig:case1}
\end{figure}

\begin{figure}[ht!]
    \centering
    \begin{minipage}{0.99\linewidth}
        \centering
        \includegraphics[width=\linewidth]{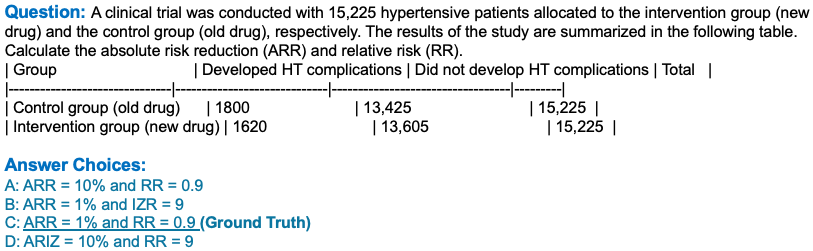}
    \end{minipage}
    \vspace{1ex}
    \\
    \begin{minipage}{0.99\linewidth}
        \centering
        \includegraphics[width=\linewidth]{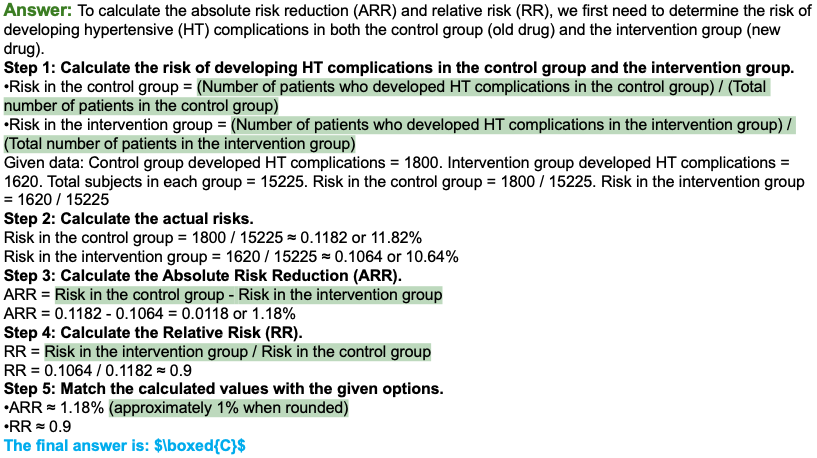}
    \end{minipage}
    \caption{Question and answer pair for Case 2. 
    \textcolor{cyan}{\textbf{Cyan text}} highlights the final predicted choices. 
    \colorbox{clinicalgreen}{\textcolor{black}{Green highlight}} are used to emphasize reasoning steps and key clinically key information.
    }
    \label{fig:case2}
\end{figure}

\begin{figure}
    \centering
    \includegraphics[width=0.99\linewidth]{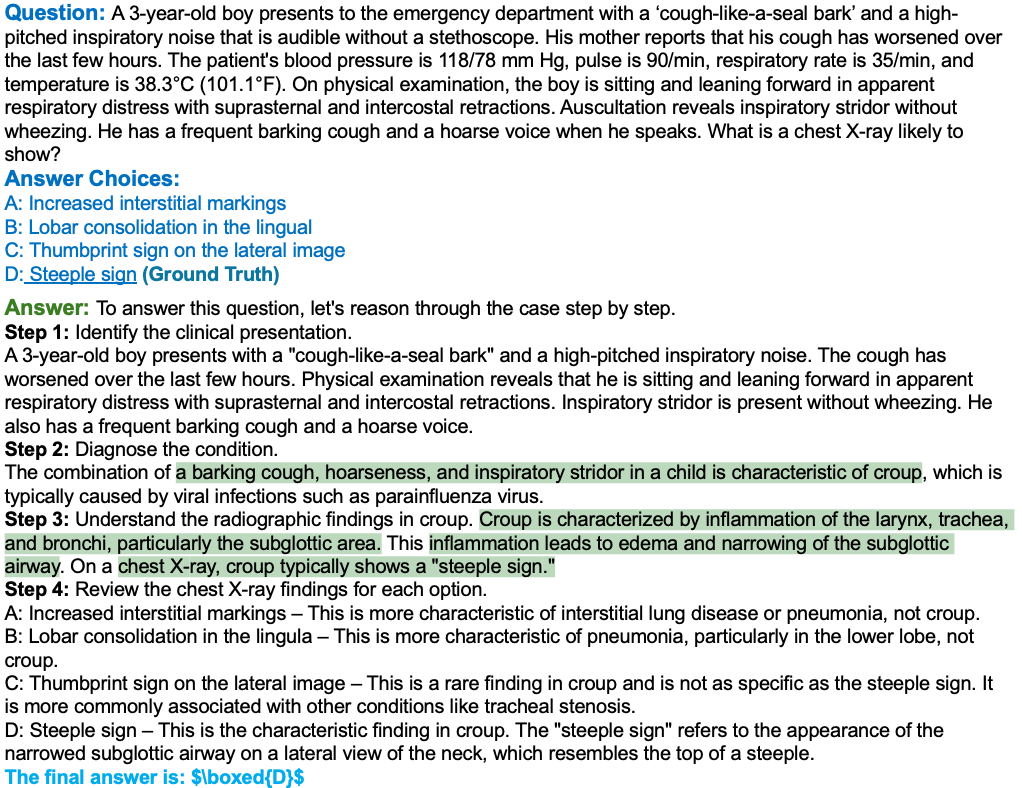}
    \caption{Question and answer pair for Case 3. 
    \textcolor{cyan}{\textbf{Cyan text}} highlights the final predicted choices. 
    \colorbox{clinicalgreen}{\textcolor{black}{Green highlight}} are used to emphasize reasoning steps and key clinically key information.
    }
    \label{fig:case3}
\end{figure}

We present three examples predicted by our model trained with minimalist RL, demonstrating interpretable step by step clinical reasoning across diverse case types. In Fig.~\ref{fig:case1}, the model correctly identifies inappropriate and potentially harmful options (e.g., use of NOACs in patients with mechanical heart valves) and adheres to guidelines by recommending bridging strategies based on patient risk factors and procedural context. In Fig.~\ref{fig:case2}, it performs multi step numerical reasoning to derive absolute risk reduction (ARR) and relative risk (RR), showcasing its ability to integrate clinical knowledge with quantitative interpretation. In Fig.~\ref{fig:case3}, the model applies structured reasoning to diagnose croup in a pediatric patient, identifying clinical features, linking them to pathophysiology, and reviewing radiographic findings, despite being supervised only on the final answer choice. This highlights the model’s capacity for guideline aligned reasoning and emergent interpretability, even without supervision on intermediate reasoning traces.

%%%%%%%%%%%%%%%%%%%%%%%%%%%%%%%%%%%%%%%%%%%%%%%%%%%%%%%%%%%%
\clearpage
\newpage

\end{document}